\documentclass{article}

\usepackage{microtype}
\usepackage{graphicx}
\usepackage{subfigure}
\usepackage{booktabs} 

\usepackage{hyperref}

\usepackage[arxiv]{icml2025} %

\usepackage{amsmath}
\usepackage{amssymb}
\usepackage{mathtools}
\usepackage{amsthm}

\usepackage{framed}

\usepackage{bm}
\newcommand{\vect}[1]{\mathbf{#1}}
\newcommand{\xm}{\vect{w}}
\newcommand{\xs}{\vect{w}^*}
\newcommand{\hess}{\vect{H_{\mathcal{L}}}}

\usepackage{arydshln}
\usepackage{color}

\usepackage[capitalize,noabbrev]{cleveref}

\usepackage{pifont}
\newcommand{\cmark}{\ding{51}}%
\newcommand{\xmark}{\ding{55}}%

\theoremstyle{plain}

\theoremstyle{definition}

\theoremstyle{remark}

\usepackage[textsize=tiny]{todonotes}

\icmltitlerunning{EfficientLLM}

\begin{document}

\twocolumn[
\icmltitle{EfficientLLM: Scalable Pruning-Aware Pretraining for Architecture-Agnostic Edge Language Models}



\icmlsetsymbol{equal}{*}

\begin{icmlauthorlist}
\icmlauthor{Xingrun Xing}{xxx,yyy}
\icmlauthor{Zheng Liu}{yyy}
\icmlauthor{Shitao Xiao}{yyy}
\icmlauthor{Boyan Gao}{zzz}
\icmlauthor{Yiming Liang}{xxx,yyy}
\icmlauthor{Wanpeng Zhang}{zzz2}
\icmlauthor{Haokun Lin}{xxx}
\icmlauthor{Guoqi Li}{xxx}
\icmlauthor{Jiajun Zhang}{xxx}
\end{icmlauthorlist}

\icmlaffiliation{xxx}{Institute of Automation, Chinese Academy of Sciences}
\icmlaffiliation{yyy}{Beijing Academy of Artificial Intelligence}
\icmlaffiliation{zzz}{University of Oxford}
\icmlaffiliation{zzz2}{Peking University}

\icmlcorrespondingauthor{Zheng Liu}{zhengliu1026@gmail.com}
\icmlcorrespondingauthor{Jiajun Zhang}{jjzhang@nlpr.ia.ac.cn}

\icmlkeywords{Machine Learning, ICML}

\vskip 0.3in
]



\printAffiliationsAndNotice{}  

\begin{abstract}
Modern large language models (LLMs) driven by scaling laws, achieve intelligence emergency in large model sizes. Recently, the increasing concerns about cloud costs, latency and privacy make it an urgent requirement to develop compact edge language models. Distinguished from direct pretraining that bounded by the scaling law, this work proposes the pruning-aware pretraining, focusing on retaining performance of much larger optimized models. It features following characteristics: 1) Data-scalable: we introduce minimal parameter groups in LLM and continuously optimize structural pruning, extending post-training pruning methods like LLM-Pruner and SparseGPT into the pretraining phase. 2) Architecture-agnostic: the LLM architecture is auto-designed using saliency-driven pruning, which is the first time to exceed SoTA human-designed LLMs in modern pretraining. We reveal that it achieves top-quality edge language models, termed EfficientLLM, by scaling up LLM compression and extending its boundary. EfficientLLM significantly outperforms SoTA baselines with $100M \sim 1B$ parameters, such as MobileLLM, SmolLM, Qwen2.5-0.5B, OLMo-1B, Llama3.2-1B in common sense benchmarks. As the first attempt, EfficientLLM bridges the performance gap between traditional LLM compression and direct pretraining methods, and we will fully open source at \url{https://github.com/Xingrun-Xing2/EfficientLLM}.
\end{abstract}

\section{Introduction}

Large Language Models (LLMs) have become a central component of modern AI systems \cite{achiam2023gpt,guo2025deepseek} and are increasingly transforming daily life, particularly in mobile edge appilications.
However, typical LLMs \cite{touvron2023llama}, with 7 billion to 1 trillion parameters, require on-cloud deployment and continuous internet connectivity for interface. This places significant challenges in terms of latency, data-security and cloud-costs. In fact, fully using LLMs for mobile edge applications can be impractical, which requires approximate one million H100 GPUs \cite{liu2024mobilellm}.
As a result, developing edge language models on resource-constrained devices becomes a recent tendency. 
For instance, MobileLLM \cite{liu2024mobilellm} focuses on sub-one billion model sizes, which would fit in the DRAM of smartphones without excessive consumption.

Direct pretraining is dominant in recent tiny language model pretraining. 
Some practices such as MobileLLM and PanGu-$\pi$-Pro \cite{tang2024rethinking} design deep-and-thin architectures for model efficiency. Other practices such as TinyLlama \cite{zhang2024tinyllama} and Qwen2.5-0.5B \cite{yang2024qwen2} focus on scaling up pretraining data to 3T and 17T tokens.
Based on best architectures and sufficient data, modern tiny models \cite{yang2024qwen2,groeneveld2024olmo} are encouraging to touch the performance boundary.
However, their overall performance appears to somewhat locked by the parameter scaling law \cite{kaplan2020scaling}: given limited model size, simply scaling up pretraining data is inefficient. More importantly, the intelligence emergency \cite{brown2020language} is only observed on larger model sizes, meaning tiny models may never acheve this by direct pretraining alone. What is the next to train more efficient edge models remains an open challenge.

\begin{figure*}[t]
\begin{center}
\centerline{\includegraphics[width=1.0\textwidth]{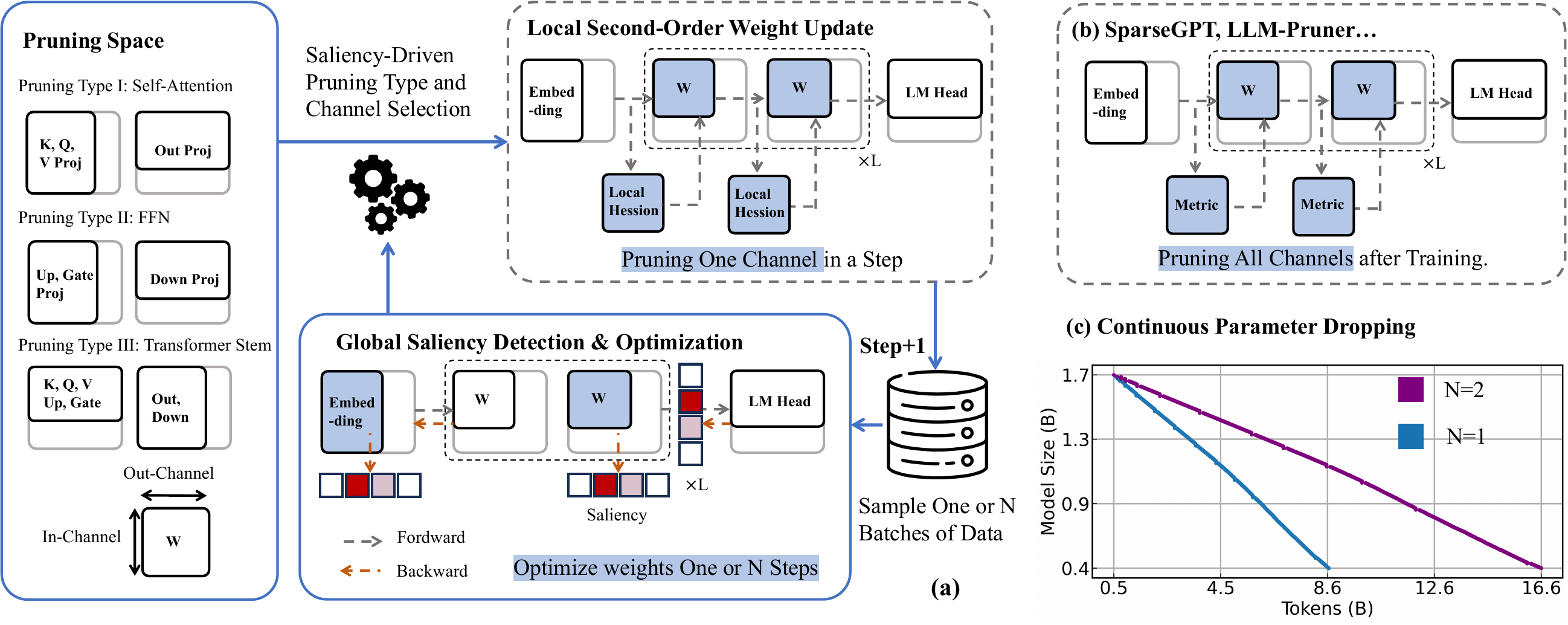}}
\vskip -0.1in 
\caption{
An overview of pruning-aware pretraining. (a) Training loop includes the joint saliency detection and weight optimizing, pruning type selection from pruning space, and second-order weight updating. (b) Traditional post-training pruning can be embeded in the training loop to scale up. (c) Continuous model size compression in pretraining.}
\label{f2}
\end{center}
\vskip -0.3in 
\end{figure*}

In parallel, LLM compression \cite{ashkboos2024slicegpt, gu2024minillm,ashkboos2024quarot} focuses on retaining the performance of larger and stronger models while reducing computational cost. 
Despite its protential efficiency, existing methods \cite{sreenivas2024llm,frantar2023sparsegpt, xiao2023smoothquant} compress LLM only using a small calibration dataset in post-training, which often results in significant performance degradation, making them unsuitable for top-quality edge language models.
Recently, ShearedLlama \cite{xia2023sheared} initializes from an optimized LLM, improving training efficiency. However, the constrained optimization \cite{platt1987constrained} hinders scaling up pruning stage and the performance gap to direct pretraining still remains. This work extends the performance boundary of traditional LLM compression by scaling up training data, which is underexplored but essential in this field.

\begin{figure}[t]
\begin{center}
\centerline{\includegraphics[width=0.99\columnwidth]{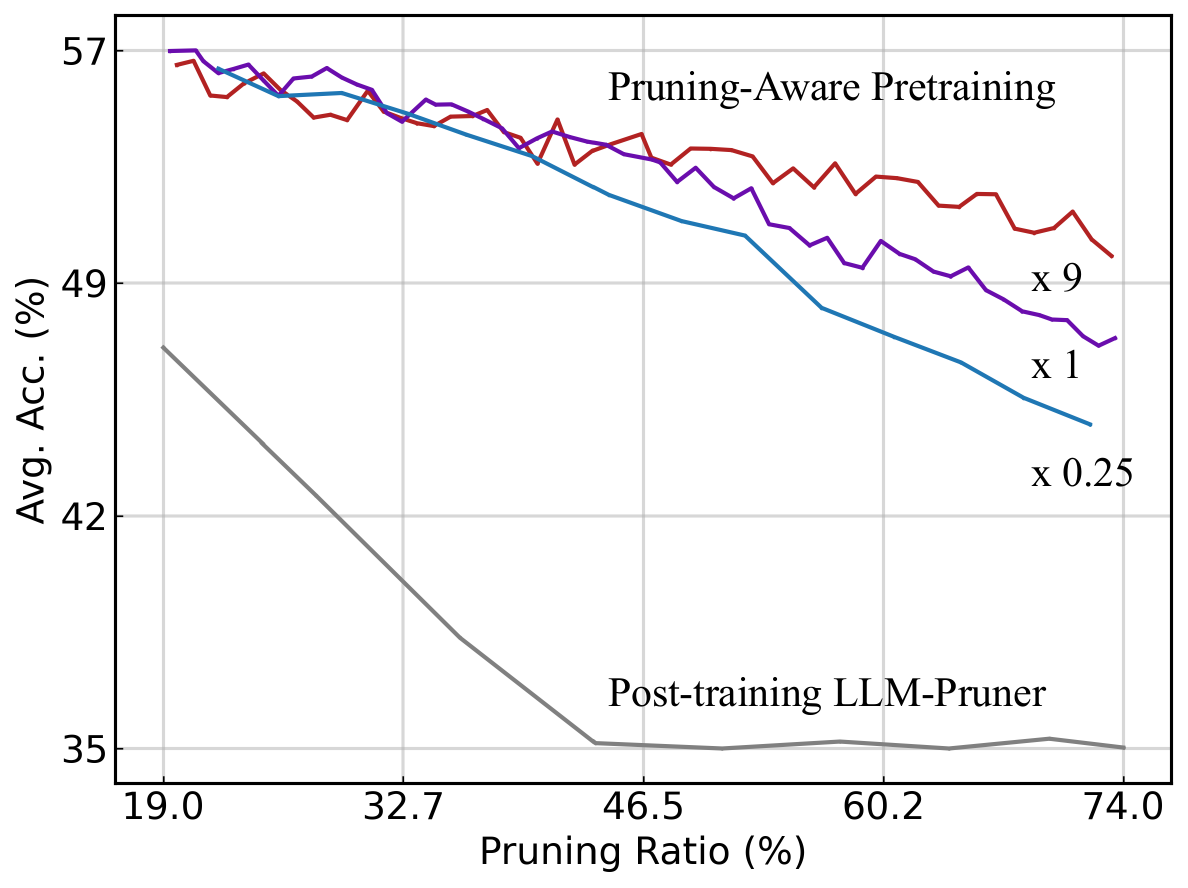}}
\vskip -0.1in
\caption{Performance of Pruning-Aware Pretraining. By scaling up LLM-Pruner in pretraining, performance  of the source model is retained even if the pruning rate more than 70\%.}
\label{f1}
\end{center}
\vskip -0.3in 
\end{figure}

This work proposes the pruning-aware pretraining to extend the efficiency boundary of edge language models. 
A family of top-efficiency edge language models in $100M \sim 1B$ sizes are pretrained, named EfficientLLM. 
As shown in Fig. \ref{f2}, we fomulate pruning-aware pretraining as a bi-level optimization problem, and decouple the LLM pruning at every pretraining step. 
Driven by saliency, the overall architecture can be auto-designed \cite{zoph2018learning,yu2020bignas} according to predefined pruning space step by step.
Compared with direct pretraining, pruning-aware pretraining leverages the performance of much larger optimized models, which direct pretraining smaller models never achieves. 
Compared with post-training pruning, it scales up the pruning stage with pretraining data. 
As shown in Fig. \ref{f1}, pruning-aware pretraining scales up vanilla LLM-Pruner, achieving more than a 10\% increase in accuracy. 
This work advances both edge language models and LLM compression:
\begin{itemize}
\item We propose a family of SoTA edge language models, EfficientLLM, in $100M \sim 1B$ sizes. EfficientLLM exceeds traditional parameter scaling law and extends efficiency boundary of tiny models.
\item We propose the pruning-aware pretraining, promating LLM compression to the era of pretraining. By scaling up training data, vanilla LLM-Pruner significantly exceeds SoTA methods without bells and whistles.
\item We explore the auto-designd architectures in modern pretraining for the first time. Saliency-driven architectures are competitive with human best practices.
\end{itemize}

\section{Preliminary and Related Works}

\textbf{Edge Language Models.} Modern large language models are powered by the scaling law \cite{kaplan2020scaling}: larger models achieve higher data efficiency, making optimal training favor large models with moderate data.
Towards accurate compact models, a lot of efforts explore the optimal training recipes: 1) data scale. OLMo-1B \cite{groeneveld2024olmo}, TinyLlama-1.1B \cite{zhang2024tinyllama}, Qwen2.5-0.5B \cite{yang2024qwen2} pretrain on 2T, 3T, and 17T tokens respectively, which is significantly larger than the optimal data sizes according to scaling law. 2) Architectures. MobileLLM \cite{liu2024mobilellm} shows that the deep-and-thin network and layer sharing achieve additional performance gains. 
However, previous works are bounded by the scaling law, and can be data-inefficient in pretraining. 
More recently, Llama3.2 \cite{dubey2024llama} and MiniTron \cite{sreenivas2024llm} introduce distillation and pruning for data-efficient training. There are mainly 2 drawbacks which addressed in this work: 1) the LLM pruning itself does not scale up. MiniTron only uses a small calibration dataset for pruning and only scales up recovery training, while this work scales up pruning itself to retain more performance. 2) the distillation \cite{ko2024distillm} in pretraining is not training-efficient. Teacher models are typical 7B sized LLM \cite{touvron2023llama2} with more than $\times 50$ FLOPs than a sub-billion edge model, which we delate in EfficientLLM.

\textbf{LLM Pruning} \cite{dong2024pruner,zhang2024plug,zhao2024apt,bhaskar2024finding}. We mainly focus on structural pruning \cite{chen2023lorashear,choukse2018compresso} for hardware friendly edge models. The most widely used LLM pruning is based on the Taylor expansion \cite{lecun1989optimal,hassibi1993optimal,van2023llm}, as shown in Table \ref{t1}. By calibration, typical SparseGPT \cite{frantar2023sparsegpt} and Wanda \cite{sun2023simple} can only applied in simi-structured pruning; LLM-Pruner \cite{ma2023llm} only achieves 20\% pruning ratio with reasonable accuracy. Even if pruning with finetuning, LoraPrune \cite{zhang2023loraprune} can only prune in 50\% ratio. So there is an urgent requirement to scale up LLM pruning in pretraining.
Another line of works learn to initialize from source model such as ShearedLlama \cite{xia2023sheared} and NutePrune \cite{li2024nuteprune} with less than 0.5B tokens. 
However the constraint optimization is biased in large-scale pretraining and this work scales up Taylor expansion metrics.

\section{Pruning-Aware Pretraining}

According to scaling laws, both the scale of training data and the number of parameters are fundamental to the emergence of intelligence in modern LLMs. Direct pretraining of smaller models is inefficient and lacks generalization ability. Model compression methods, although based on pretrained large models, fail to meet the data scale requirements and suffer from significant performance drop.

The principle of this work is to bridge the gap between direct pretraining and LLM compression by condersiding both the source model scale and pretraining data scale.
In practice, pruning-aware pretraining continuously drops insignificant parameters and scales up pruning in pretraining. 

\textbf{Problem Formulation.} Finding a sub-network from a pretrained LLM is non-trivial. Given an optimized LLM, post-training LLM pruning focuses on finding optimal channels in each layer towards a target architecture. However,
for edge language models, it is still challenging to define the efficient target architecture from its source model. For instance, MobileLLM shows the deeper architecture is better than the wider for sub-billion LLMs by human design and practice. This best practice can be sub-optimal in the context of the given source model, because each source model has distinguished salient pruning target. 
We fomulate the architecture-agnostic pruning problem as:
\begin{equation}
\label{e1}
\underset{a \in \mathcal{A}}{\text{min }} \underset{c \in \mathcal{C}}{\text{min }} \underset{w}{\text{min }} \mathcal{L}_{pretrain}(a, c, w|\mathcal{M}),
\end{equation}
where $\mathcal{A}$ and $\mathcal{C}$ are sub-architectures and sub-channels sampled from the source model $\mathcal{M}$. We jointly optimize pretraining loss through three factors: 1) the sub-architecture \cite{cai2019once}, 2) the sub-channels, and 3) the model weights. 
We outline the pruning-aware pretraining in Fig. \ref{f2} and detail each part in the following subsections.

\begin{table}[t]
    \centering
    \caption{Comparison between Taylor-expansion based LLM pruning. ``PT", ``FT", ``CB" denote pruning in the pretraining, finetuning, calibration stage respectively, and ``G", ``U", ``Ratio" denote global pruning, weight update, pruning ratio. }\label{t1}
    \renewcommand{\arraystretch}{1.02}
    \setlength{\tabcolsep}{4.5pt}
    \begin{tabular}{l | c c c | c c c}
        \toprule
        \textbf{Method}  & \textbf{PT} & \textbf{FT} & \textbf{CB} & \textbf{G}  & \textbf{U} & \textbf{Ratio} \\
        \midrule
        Magnitude &  -- &  -- &  -- &  \xmark &  \xmark & --   \\
        SparseGPT &  &   & \cmark   &  \xmark &  \cmark & 2:4  \\
        Wanda     &  &    & \cmark  &  \xmark &  \xmark & 2:4  \\
        LLM-Pruner& &    & \cmark   &  \cmark &  \xmark & 20\%  \\
        LoraPrune &  & \cmark   &   &  \cmark &  \xmark & 50\%  \\
        EfficientLLM-A & \cmark &   & &  \cmark &  \xmark & $>$70\%  \\
        EfficientLLM-B & \cmark &   & &  \cmark &  \cmark & $>$70\%  \\
        \bottomrule
    \end{tabular}
    \vskip -0.15in
\end{table}

\subsection{Defining Minimal Pruning Group}

\begin{figure*}[t]
\begin{center}
\centerline{\includegraphics[width=1.0\textwidth]{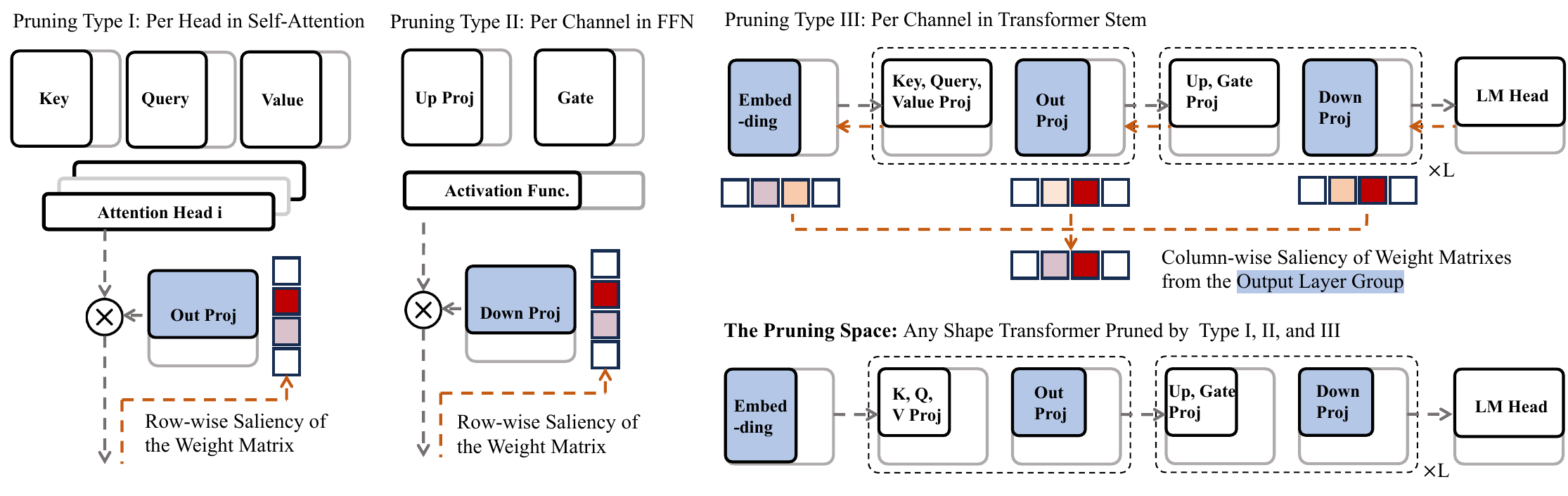}}
\vskip -0.1in
\caption{
    Three basic pruning typies in the pruning space. We plot all the weight metrics with shape $[D_{input}, D_{output}]$.
    In backpropagation (in orange), the saliency of the output layer group (in blue) is calculated according to Eq. \ref{e9}.
}
\label{f3}
\end{center}
\vskip -0.3in
\end{figure*}

To achieve architecture-agnostic pruning, we first define the minimal parameter groups as the minimal architectures of the network, which should be flexable enough to construct any shape transformer models. Given an optimized large model $\mathcal{M}$, the pruned model $\mathcal{M^*}$ can be represented as:
\begin{equation}
\label{e2}
\mathcal{M^*} = \mathcal{M} - \sum\limits_{t=1}^{n} g_t,  \quad \text{s.t.}  \quad \underset{g_t \in \mathcal{G}}{\text{min}} \mathcal{L}_{pretrain}(\mathcal{M}),
\end{equation}
where $g_t$ is the mini-group of parameters pruned in step t, and $\mathcal{G}$ is the pruning space formulated by defined mini-groups. According to Eq. \ref{e2}, 
the pruning are decoupled by t steps and can be approximately solved sequentially:
\begin{align}
	\quad & \mathcal{M}_{t} = \mathcal{M}_{t-1} - g_t^* \label{e3} \\
	 & \text{s.t.} \quad g^*_t = \underset{g_t \in \mathcal{G}}{\mathrm{argmin}} \enskip \mathcal{L}_{pretrain}(g_t | \mathcal{M}_{t-1}). \nonumber
\end{align}
We first assumpt an optimal $g^*_t$ in each pruning step with respect to the pretraining loss, and solve how to acquire $g^*_t$ in the next subsection. In each pruning step, an optimal mini-group of parameters are selected and dropped from the pretraining LLM $\mathcal{M}_{t-1}$, allowing the model $\mathcal{M}$ to adaptively reduce the number of parameters until a specific computation budget is met.

Towards fully structured pruning, two crucial constraints are considered in the mini-group design: 
(i) coupled architectures in different layers and (ii) consistent shape \cite{xia2023sheared} in different blocks, where the former is introduced by LLM-Pruner to confirm all relevant parameters are pruned at the same time, and the later provides further speeding up in the system level. The difference from LLM-Pruner pruning space is that we split the pruned parameters into minimal groups and are able to adaptively combinate during pretraining condersiding training dynamics, instead of the handcraft target in LLM-Pruner. Additionally, we couple channels in all blocks by saliency to satisfy constraint (ii). 

For simplify, we indicate the \textit{\textbf{input layer group}} as \textit{query, key, value projections} in attention blocks; and \textit{up, gate projections} in feed forward blocks. We indicate the \textit{\textbf{output layer group}} as the \textit{output projections} in attention blocks; and \textit{down projections} in feed forward blocks.
As shown in Fig. \ref{f3}, three basic pruning types are defined: \\
(i) Per-head pruning in self-attention blocks: when an attention head is pruned, all the corresponding output channels in the input layer group and input channels in the output layer group are pruned at the same time. Additionally, we select the mini-group $\mathcal{G}^{(\ell)}_\text{attn}$ in each block with the minimal saliency in step t, and merge $\mathcal{G}^{(\ell)}_\text{attn}$ in all blocks as $\mathcal{G}_\text{attn}$:
\begin{equation}
\label{e4}
\mathcal{G}_\text{attn} = \{ W_\text{:,i:j}^{(k,\ell)},W_\text{:,i:j}^{(q,\ell)},W_\text{:,i:j}^{(v,\ell)},W_\text{i:j,:}^{(o,\ell)}, \ell=1,2,...,n \},
\end{equation}
where $W_{:,i:j}$ and $W_{i:j,:}$ are column-wise and row-wise pruned, and i:j corresponds to channels of an attention head. \\
(ii) Per-channel pruning in feed-forward blocks: when a channel is pruned in the intermediate activation, the coupled channels include one output channel in the input layer group, and one input channel in the output layer group in a FFN block. Across different blocks, we also couple the minimal-saliency groups $\mathcal{G}^{(\ell)}_\text{ffn}$ and merge to $\mathcal{G}_\text{ffn}$:
\begin{equation}
\label{e5}
\mathcal{G}_\text{ffn} = \{ W_\text{:,i}^{(up,\ell)},W_\text{:,i}^{(q,\ell)},W_\text{i,:}^{(down,\ell)}, \ell=1,2,...,n \}.
\end{equation}
(iii) Per-channel pruning in the transformer stem: when a channel of the transformer stem is pruned, as shown in Fig.\ref{f3}, one coupled channel in the token embedding, one input channel in input layer group and one output channel in output layer group for every block, one input channel of the LM head projection is correspondingly pruned at the same time. We donate the stem mini-group as $\mathcal{G}_\text{stem}$:
\begin{align}
\mathcal{G}_\text{stem} = & \{ W_\text{i,:}^{(k,\ell)},W_\text{i,:}^{(q,\ell)},W_\text{i,:}^{(v,\ell)},W_\text{:,i}^{(o,\ell)}\},... \label{e6} \\
& \cup \{ W_\text{i,:}^{(up,\ell)},W_\text{i,:}^{(gate,\ell)},W_\text{:,i}^{(down,\ell)}\},... \nonumber\\
& \cup \{ \mathbf{w}_\text{i}^{(emb)},W_\text{i,:}^{(head)} \}, \quad \ell=1,2,...,n \nonumber
\end{align}
where the i should be the same in every blocks, while i,j needn't the same across blocks in Eq.\ref{e4},\ref{e5}.

Given a transformer with hidden size $m$, head number $h$, intermediate size $n$, and $l$ layers, the original pruning space is $h^{(\ell)} \times n^{(\ell)} \times m$.
Notice that, the mini-groups are dynamically grouped by saliency in each pruning step, and we only choose among the 3 types to prune in step t.
By coupling the parameters into mini-groups, the choice space is reduced to 3 in each step of Eq.\ref{e3}, and the final pruning space is 3$^t$.

\subsection{Optimizing Mini-Groups by Saliency}
Based on the mini-groups, Eq.\ref{e1} becomes a bi-level optimization problem of the mini-groups $g$ and weights $w$:
\begin{align}
\quad & \underset{g \in \mathcal{G}}{\text{min }} \mathcal{L}_{pretrain}(g, w^*|\mathcal{M}), \label{e7}\\
& \text{s.t.} \quad w^* = \underset{w}{\mathrm{argmin}} \enskip \mathcal{L}_{pretrain}(w, g^* | \mathcal{M}), \nonumber
\end{align}
where the outer optimization could be solved by Eq.\ref{e3}, and the inner optimization could be directly solved by gradient descent. 
In pretraining, gradient descent and mini-group optimization (Eq.\ref{e3}) alternate. We refer to the alternating approach of one step of gradient descent followed by one step of pruning as pruning-aware pretraining $\times$1.

In the mini-group optimization step, Taylor expansion evaluates the optimal mini-group $g^*_t$ in Eq.\ref{e3}. 
For an optimized model, loss of any weight $\xm$ can be approximated by a second-order Taylor expansion around its optimal value $\xm^*$:
\begin{equation}
  \label{e8}
  \mathcal{L}(\xm) \simeq \mathcal{L}(\xm^*) + \delta\mathbf{w}^\top \nabla \mathcal{L}(\xs) + \frac{1}{2} \delta\mathbf{w}^\top \hess (\xs) \delta\mathbf{w}
\end{equation}
where $\mathcal{L}$, $\nabla \mathcal{L}$, $\hess$ is the global loss, gradient, hession matrix; and $\delta\mathbf{w}=\mathbf{w}-\mathbf{\xs}$. We substitute Eq.\ref{e8} into Eq.\ref{e3}:
\begin{align}
    g^*_t &= \underset{g_t \in \mathcal{G}}{\mathrm{argmin}} \enskip \mathcal{L}_{pretrain}(g_t | \mathcal{M}_{t-1}) \label{e9} \\
    &= \underset{g_t \in \mathcal{G}}{\mathrm{argmin}} \enskip    g_t^\top \nabla \mathcal{L}(\mathcal{M}_{t-1}) + \frac{1}{2} g_t^\top \hess (\mathcal{M}_{t-1}) g_t, \nonumber
\end{align}
where we omit the first term $\mathcal{L}(\xm^*) = \mathcal{L}(\mathcal{M}_{t-1})$ in Eq.\ref{e8}, because $\mathcal{L}(\mathcal{M}_{t-1})$ is constant for the 3 mini-groups, $\mathcal{G}=\{ \mathcal{G}_\text{attn},\mathcal{G}_\text{ffn},\mathcal{G}_\text{stem} \}$.

As shown in Fig.\ref{f3}, we define the calculation of mini-group saliency based on Eq.\ref{e9}, and then couple the mini-groups. \\
\textit{Pruning Type I}: in each attention block, we only calculate element-wise saliency matrix for the output projection.
To evaluate input channels, the saliency matrix is summed row-wise.
We select $\mathcal{G}_\text{attn}^{(\ell)}$ with the minimal row-wise saliency in each block, and then couple to $\mathcal{G}_\text{attn}$.
The group saliency $\mathcal{S}_\text{attn}$ is the summation over $\mathcal{G}_\text{attn}$. \\
\textit{Pruning Type II}: in each FFN, we only calculate element-wise saliency matrix for the down projection as Eq.\ref{e9}, and then sum row-wise. $\mathcal{G}_\text{ffn}^{(\ell)}$ with the minimal summed saliency are similarily coupled as $\mathcal{G}_\text{ffn}$ with saliency $\mathcal{S}_\text{ffn}$. \\
\textit{Pruning Type III}: in stem, we already have all the element-wise saliency in the output layer group based on Type I, II.  
To evaluate input channels, saliency matrixes are summed column-wise in a layer.
We then sum column-wise saliency over the output layer group and select the minimum as $\mathcal{S}_\text{stem}$. \\
In each step, the optimization in Eq.\ref{e3} is finally solved by:
\begin{equation}
    g^*_t = \underset{g_t \in \mathcal{G}}{\mathrm{argmin}} \enskip \{\mathcal{S}_\text{attn}, \mathcal{S}_\text{ffn}, \mathcal{S}_\text{stem} \}. \label{e10}
\end{equation}

\textbf{Efficient Saliency Calculation.} In pretraining, we speed up the saliency calculation from 2 aspects: 1) Output layer group only calculation: a neural network is a directed acyclic graph (DAG) \cite{liu2018darts}. For each node in the graph, pruning all of its inputs or all of its outputs is sufficient to prune the entire network. 
Additionally, since the number of parameters in the input layer group of a Transformer is 2 $\sim$ 3 times that of the output layer group, we only calculate the saliency of the output layer group for pruning. \\
2) Approximate Hession matrix: 
existing works such as LLM-Pruner, SparseGPT, and Wanda have proposed a series of Hession approximations to speed up. By substituting Eq.\ref{e9}, this framework is general to cooperate with previous post-training pruning metrics. Without loss of generalization, we choose LLM-Pruner to scale up.

\subsection{Second-Order Weight Updating}
Existing second order pruning applies the same Hession matrix for the pruning weight detection and the remaining weight updating. However, calculating the global Hession matrix is impossible in modern LLMs for its $\mathcal{O}(n^4)$ complexity. A common approach is to use the squared error at each layer as a proxy for the global loss: $\hess \simeq XX^T$, such as in SparseGPT, OBC \cite{frantar2022optimal}. Although achieving the $\mathcal{O}(d_{row} \times d_{col}^2)$ complexity, Hessian matrixes can not capture the global loss. 

This work solves this problem by decouple the Hession matrix in saliency detection and weight updating. In saliency detection, we approximate with global diagonal Hession matrixes as LLM-Pruner to detect global saliency; in weight updating, we apply the layerwise approximation as $\hess \simeq XX^T$ to minimize pruning error. And the remaining weights can be updated by ${\delta}w_p = - \frac{w_p}{[\mathbf{H}^{-1}]_{pp}} \cdot \mathbf{H}^{-1}_{:, p}$.
Notice that, in each pruning, we only prune a mini-group including only one colomn of weights in a layer. To efficiently calculate the inverse of Hession matrix $\mathbf{H}^{-1}_{:, p}$, we only need to solve a linear equation: $\mathbf{e_p} = \mathbf{H}\mathbf{H}^{-1}_{:, p}$ in a step.

\textbf{Discussion.} This work advances LLM pruning in 3 aspects: 
1) scaling up LLM pruning in pretraining. Althrough some industrial LLMs such LlaMA-3.2 and MiniTron are also pretrained from larger models, the pruning stage itself may not scale up. MiniTron only iteratively prunes 4 times by a small calibration dataset, while the pruning-aware pretraining is continuously optimized by large-scale pretraining data. 
2) Target-agnostic pruning. The auto-designed architectures \cite{xu2021bert,wu2019fbnet} achieve competitive results to SoTA human designed LLMs in modern pretraining for the first time.
3) Efficient Second-Order Updating. We propose efficient Hession approximations, making the second-order updating acceptable in pretraining.

\begin{table*}[t]
\renewcommand\arraystretch{0.6}
\centering
\caption{Zero-shot performance on World Knowledge and Common Sense Reasoning tasks. ``Avg.'' calculate among the 7 Common Sense Reasoning tasks. \#Tokens count continued pretraining for EfficientLLM. All the results are evaluated on the same evaluation (Appendix B), except MobileLLM \cite{liu2024mobilellm}, because its close source evaluation can not be reproducted and we cite from the paper.}
\label{t2}
\setlength{\tabcolsep}{1.5mm}
{\resizebox{\textwidth}{!}{
\begin{tabular}{lccc cccc cccc}
\noalign{\vspace{0.1em}}\hline\noalign{\vspace{0.1em}}
\noalign{\vspace{0.1em}}\hline\noalign{\vspace{0.1em}}
\textbf{Model} & \textbf{\#Tokens} & \textbf{\#Params} & \textbf{MMLU} & \textbf{ARC-c} & \textbf{ARC-e} & \textbf{BoolQ} & \textbf{HellaSwag} & \textbf{OBQA} & \textbf{PIQA} & \textbf{WinoGrande} & \textbf{Avg.} \\ 
\noalign{\vspace{0.1em}}\hline\noalign{\vspace{0.1em}}
OPT-125M & 180B & 125M & 26.02  & 22.87 & 43.31 & 55.44 & 31.37 & 27.80 & 62.62 & 49.80 & 41.89 \\
GPT-neo-125M & 300B & 125M & 26.89  & 23.29 & 43.22 & 61.77 & 30.49 & 26.00 & 62.62 & 51.93 & 42.76 \\
Pythia-160M & 300B & 162M &  26.43 & 22.27 & 37.84 & 43.33 & 29.97 & 26.40 & 58.87 & 49.96 & 38.38 \\
Memba-130M & 1.2T & 130M & 27.65  & 24.49 & 47.56 & 54.68 & 35.11 & 29.00 & 64.69 & 53.35 & 44.13 \\
MobileLLM-LS-125M & 1T & 125M & -- & 28.7 & 45.8 & 60.4 & 39.5 & 41.1 & 65.7 & 52.1 & 47.61 \\
SmolLM-135M & 600B & 135M &  30.05  & 29.35 & 61.32 & 59.85 & 42.67 & 34.40 & 68.55 & 52.96 & 49.87 \\
EfficientLLM-A & 500B & 134M & 30.54  & 30.97 & 62.88 & 60.40 & 43.81 & 33.60 & 68.82 & 53.28 & 50.54 \\
\noalign{\vspace{0.1em}}\hdashline[0.8pt/1pt]\noalign{\vspace{0.2em}} 
OPT-350M & 180B & 331M &  26.96 & 23.98 & 44.02 & 57.80 & 36.63 & 27.80 & 64.91 & 52.96 & 44.01 \\
BLOOM-560M & 350B & 559M &  27.32 & 24.40 & 46.04 & 44.46 & 36.54 & 28.80 & 62.57 & 53.20 & 42.29 \\
Pythia-410M & 300B & 405M & 29.10  & 24.15 & 51.39 & 59.20 & 40.20 & 29.40 & 66.70 & 53.83 & 46.41 \\
MobileLLM-LS-350M & 1T & 345M & -- & 32.5 & 54.4 & 62.8 & 50.6 & 45.8 & 69.8 & 57.2 & 53.30 \\
SmolLM-360M & 600B & 362M &  33.89 & 36.26 & 70.16 & 55.23 & 53.51 & 37.60 & 71.38 & 57.22 & 54.48 \\
Qwen2-0.5B & 15T & 494M &  31.85 & 28.50 & 55.05 & 61.25 & 49.16 & 32.80 & 69.75 & 57.22 & 50.53 \\
Qwen2.5-0.5B & 17T & 494M &  33.37 & 32.17 & 64.44 & 61.99 & 52.09 & 35.20 & 70.29 & 56.20 & 53.20  \\
EfficientLLM-A & 50B & 469M &  33.09 & 35.92 & 70.50 & 59.85 & 53.16 & 35.00 & 72.69 & 56.27 & 54.77 \\
EfficientLLM-A & 500B & 469M & 34.54  & 38.40 & 72.10 & 62.42 & 56.84 & 40.40 & 73.83 & 57.46 & 57.35  \\
\noalign{\vspace{0.1em}}\hdashline[0.8pt/1pt]\noalign{\vspace{0.2em}} 
OPT-1.3B & 180B & 1.3B & 29.57  & 30.03 & 57.49 & 56.54 & 53.66 & 32.80 & 72.31 & 59.04 & 51.70 \\
GPT-neo-1.3B & 380B & 1.3B & 30.00  & 25.94 & 56.31 & 61.90 & 48.99 & 33.40 & 71.00 & 54.62 & 50.31 \\
BLOOM-1.1B & 350B & 1.1B & 29.16  & 25.77 & 51.73 & 59.51 & 43.11 & 29.60 & 67.30 & 54.62 & 47.38 \\
Pythia-1B & 300B & 1.0B & 30.14  & 26.96 & 56.86 & 60.04 & 47.15 & 31.20 & 70.29 & 52.88 & 49.34 \\
TinyLlama-1.1B & 3T & 1.1B & 32.30  & 30.29 & 60.40 & 56.85 & 59.13 & 35.80 & 73.07 & 59.04 & 53.51 \\
ShearedLlama-1.3B & 50B & 1.3B & 31.51  & 29.44 & 61.07 & 61.83 & 59.33 & 34.40 & 73.94 & 58.01 & 54.00 \\
OLMo-1B & 2T & 1.2B &  32.03 & 30.72 & 63.55 & 61.38 & 62.86 & 36.40 & 75.35 & 59.35 & 55.66 \\
Llama3.2-1B & -- & 1.2B &  36.31 & 31.48 & 65.28 & 63.88 & 63.69 & 37.40 & 74.59 & 60.54 & 56.69 \\
EfficientLLM-A & 50B & 1.1B & 36.71 & 40.36 & 73.61 & 62.39 & 60.24 & 40.20 & 75.19 & 61.25 & 59.03 \\
EfficientLLM-A & 320B & 1.1B & 37.71  & 42.24 & 73.48 & 67.09 & 64.09 & 41.80 & 75.41 & 61.17 & 60.75 \\
\noalign{\vspace{0.1em}}\hline\noalign{\vspace{0.1em}}
\noalign{\vspace{0.1em}}\hline\noalign{\vspace{0.1em}}
\end{tabular}}}
\vskip -0.15in
\end{table*}

\section{Experiments}

Based on pruning-aware pretraining, we scale up LLM compression and extend the efficiency boundary of edge language models.
We pretrain a family of top-quality edge language models, named EfficientLLM.
Our results bridge the performance gap between LLM compression \cite{han2015deep} and direct pretraining methods for the first time.

\subsection{Settings}
\textbf{Models.} 
To compare with the most general post-training pruning, EfficientLLM-A basically approximates Eq.\ref{e9} as LLM-Pruner. EfficientLLM-B additionally applies the second-order weight updating based on EfficientLLM-A. Detailed settings are as follows:
1) in main results, we pretrain EfficientLLM 134M from the source model SmolLM-360M \cite{allal2024SmolLM}; EfficientLLM 469M and 1.1B from SmolLM-1.7B \cite{allal2024SmolLM}. 2) in comparisons with LLM pruning, we keep the same source models as baselines. For Llama-7B \cite{touvron2023llama}, large pruning ratios, including 50\%, 70\% are explored; 
for Llama2-7B \cite{touvron2023llama2}, the 1.3B, 2.7B target models are applied to compare with ShearedLlama. 
Architecture details are shown in Appendix A.

\textbf{Data Composition.} EfficientLLM keeps the similar data distribution with the source model: (1) in main results, our pretraining data composition is similar to SmolLM, including 220B tokens from FineWeb-Edu \cite{lozhkov2024fineweb-edu}, 28B tokens from Cosmopedia v2 \cite{benallal2024cosmopedia}, 4B tokens from Python-Edu \cite{benallal2024smollmcorpus}, and 27.5B tokens randomly sampled from OpenWebMath \cite{paster2023openwebmath}. (2) In comparisons with LLM pruning, we sample from RedPajama-1T \cite{weber2024redpajama} as pretraining data with the Llama family as the source model.

\textbf{Training.} In main results, we train EfficientLLM with both large scale pruning-aware pretraining and continued pretraining.
For EfficientLLM-134M, 460M, and 1.1B, we pretrain 50.3B, 72.1B, and 36.7B tokens for pruning-aware pretraining followed by 500B, 500B, and 320B tokens continued pretraining. Notice that, the large-scale continued pretraining is not necessary and 50B tokens also achieve competitive performance. 
Also notice that, the token number for pruning-aware pretraining is determined by iterations that reach the target number of parameters.
We typically use the batchsize of 1M tokens with $32\sim64$ A800 GPUs for $100M \sim 1.1B$ models in pretraining. 
All the training details are shown in Appendix B.

\textbf{Evalutions.} For pretrained base models, we follow Llama, MobileLLM, and ShearedLlama to evaluate Common Sense Reasoning tasks: ARC \cite{clark2018think}, BoolQ \cite{clark2019boolq}, HellaSwag \cite{zellers2019hellaswag}, OBQA \cite{mihaylov2018can}, PIQA \cite{bisk2020piqa}, and WinoGrande \cite{sakaguchi2021winogrande}. The MMLU \cite{hendrycks2020measuring} for Word Knowledge evaluation is also applied. For instruct finetuned model, we use Alpaca-Eval \cite{li2023alpacaeval} to compare the win rate.

\begin{figure}[t]
\begin{center}
\centerline{\includegraphics[width=0.99\columnwidth]{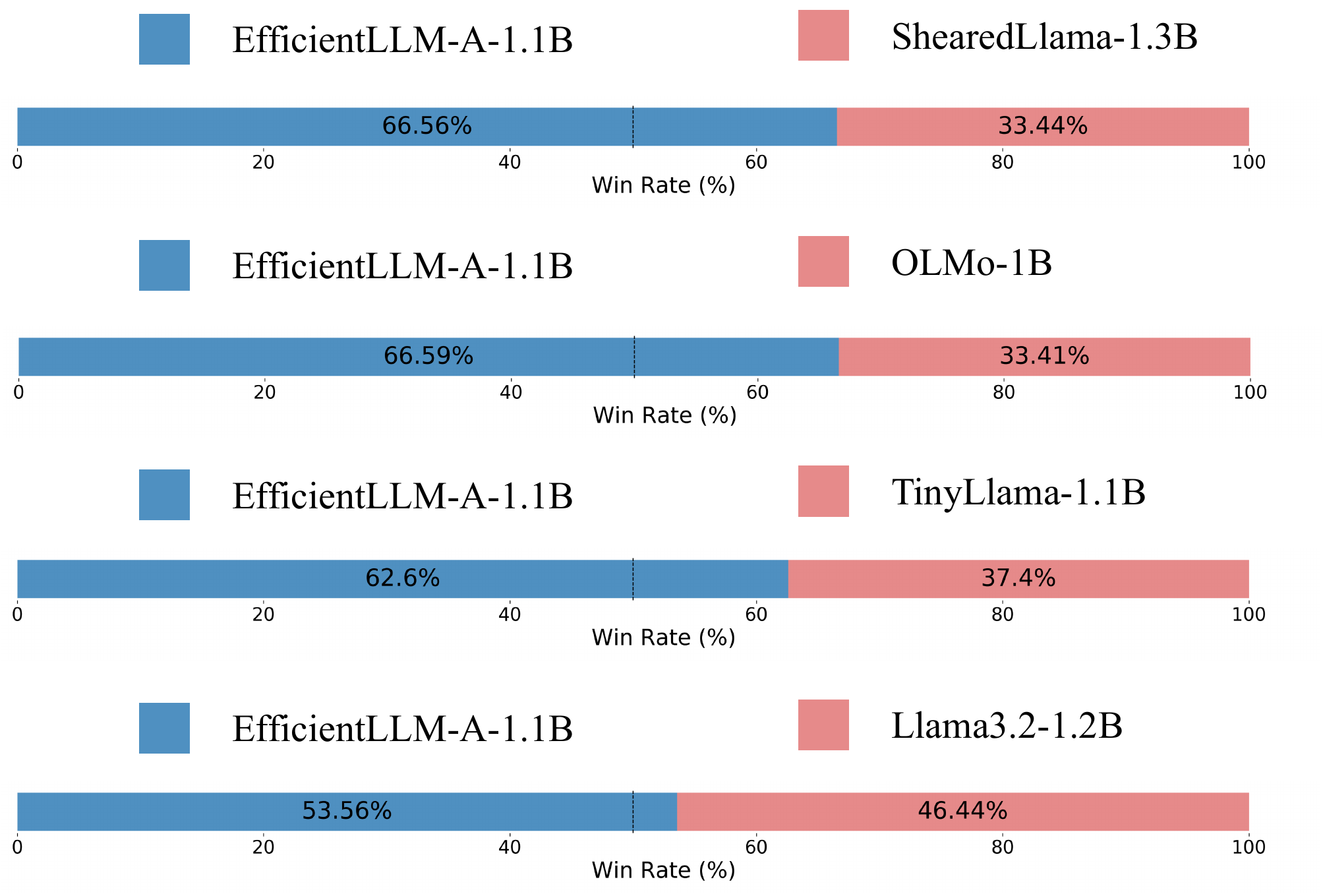}}
\caption{
    Win rate of EfficientLLM in the instruction tuning task.
}
\label{f4}
\end{center}
\vskip -0.3in 
\end{figure}

\subsection{Main Results}

\textbf{Edge Language Modeling.}
For fair comparison, we collect main streams of edge language models in $100M \sim 1B$ sizes, evaluate in the same conditions (Appendix B), and make a benchmark in Table \ref{t2}. Early edge models including OPT \cite{zhang2023opt}, GPT-neo \cite{black2022gpt}, Pythia \cite{biderman2023pythia}, and BLOOM \cite{le2023bloom} are often pretrained in limited tokens and sub-optimal architectures, which largely hinder the performance. EfficientLLM efficiently solves these problems without too much additional costs: (i) instead of direct scaling up pretraining tokens like Qwen \cite{yang2024qwen2technicalreport,yang2024qwen2}, OLMo \cite{groeneveld2024olmo}, TinyLlama \cite{zhang2024tinyllama}, EfficientLLM tries to retain the performance of larger and stronger models, which is significantly data-efficient. (ii) Through target-agnostic pruning, EfficientLLM automatically adapts to more salient architectures, achieving similar purpose in recent SoTA MobileLLM that rely on manual search for optimal architectures. As a result, EfficientLLM achieves higher accuracy even with smaller model/data sizes. For instance, EfficientLLM-134M exceeds Pythia-410M by 4.13\% average accuracy. Specifically, with the same training data, EfficientLLM-469M with 50B continued pretraining tokens exceeds SmolLM-360M with 600B tokens in Common Sense Reasoning tasks.
EfficientLLM-1.1B with 50B tokens exceeds OLMo-1B, TinyLlama, Llama3.2-1B in accuracy.
Scaling up LLM compression achieves data-efficient pretraining which exceeds the traditional LLM scaling law. 

According to Appendix B.3, most performance is retained by EfficientLLM-1.1B, and 1.05\% accuracy is dropped from SmolLM-1.7B.
Compared with SoTA industrial models such as Qwen2.5-0.5B and Llama3.2-1B trained by saturated tokens, EfficientLLM outperforms 4.15\% and 4.06\% respectively with limited pretraining data.

\textbf{Instruction Tuning.} 
We finetune EfficientLLM-1.1B base model and other 3 top-quality pretrained base models includes OLMo-1B, ShearedLlama-1.3B, TinyLlama-1.1B and Llama3.2-1B in the same condition. We use the Alpaca dataset \cite{alpaca} with 52K instructions and finetune for 3 epochs. As shown in Fig. \ref{f4}, EfficientLLM-1.1B significantly outperforms SoTA baselines, indicating the generalization ability in the supervised finetuning (SFT). 
More case studies are shown in Appendix C.

\begin{table*}[t]
\vskip -0.05in
\renewcommand\arraystretch{0.6}
\centering
\caption{Comparisons of LLM pruning in Llama-7B. We scale up pruning-aware pretraining to 5B tokens for EfficientLLM. \#Tuning donates whether to funetune after pruning. Most works report finetuned results.}
\label{t3}
\setlength{\tabcolsep}{1.5mm}
{\resizebox{\textwidth}{!}{
\begin{tabular}{clcc cccc ccc}
\noalign{\vspace{0.1em}}\hline\noalign{\vspace{0.1em}}
\noalign{\vspace{0.1em}}\hline\noalign{\vspace{0.1em}}
\textbf{\#Ratio} & \textbf{Model} & \textbf{\#Tuning} &  \textbf{ARC-c} & \textbf{ARC-e} & \textbf{BoolQ} & \textbf{HellaSwag} & \textbf{OBQA} & \textbf{PIQA} & \textbf{WinoGrande} & \textbf{Avg.} \\ 
\noalign{\vspace{0.1em}}\hline\noalign{\vspace{0.1em}}
50\% &MaP  & \cmark   & 30.63 & 49.32  & 39.69 & 42.49 & 31.40 & 66.81 & 50.67 & 44.43 \\
&MvP  & \cmark & 26.79 & 44.07 & 59.94 & 40.98 & 31.80 & 63.06 & 55.64 & 46.04 \\
&WANDA  & \cmark & 34.20 & 42.68 & 50.90 & 38.12 & 38.78 & 57.38 & 55.98 & 45.43 \\
&LLM-Pruner  & \cmark & 28.24 & 46.46 & 61.47 & 47.56 & 35.20 & 68.82 & 55.09 & 48.98 \\
&LoRAPrune  & \cmark & 31.62 & 45.13 & 61.88 & 47.86 & 34.98 & 71.53 & 55.01 & 49.72 \\
&LoRAShear  & \cmark & 32.26 & 47.68 & 62.12 & 48.01 & 34.61 & 71.80 & 56.29 & 50.40 \\
&Compresso  & \cmark & 27.82 & 48.82 & 60.09 & 39.31 & 33.40 & 66.70 & 51.93 & 46.87 \\
&NutePrune  & \xmark & 31.74 & 46.59 & 62.20 & 53.87 & 35.80 & 69.91 & 57.77 & 51.13 \\
&NutePrune  & \cmark & 32.17 & 51.68 & 62.26 & 55.88 & 34.40 & 71.00 & 57.54 & 52.13 \\
&EfficientLLM-A  & \xmark & 30.80 & 52.15 & 62.29 & 54.70 & 35.20 & 71.33 & 56.75 & 51.89 \\
&EfficientLLM-A  & \cmark & 34.04 & 64.81 & 64.83 & 60.12 & 34.60 & 73.88 & 61.48 & 56.25 \\
\noalign{\vspace{0.1em}}\hdashline[0.8pt/1pt]\noalign{\vspace{0.2em}} 
70\% & LLM-Pruner  & \cmark  & 24.83 & 39.56 & 47.28 & 31.66 & 28.80 & 60.83 & 50.75 & 40.53 \\
& NutePrune & \cmark & 26.19 & 42.17 & 62.08 & 39.43 & 30.20 & 62.30 & 51.46 & 44.83 \\
& EfficientLLM-A & \xmark & 27.73 & 54.50 & 47.89 & 47.77 & 31.00 & 68.17 & 55.17 & 47.46 \\
& EfficientLLM-A & \cmark & 29.95 & 58.59 & 58.13 & 52.02 & 34.60 & 70.08 & 55.96 & 51.33 \\
\noalign{\vspace{0.1em}}\hline\noalign{\vspace{0.1em}}
\noalign{\vspace{0.1em}}\hline\noalign{\vspace{0.1em}}
\end{tabular}
}}
\vskip -0.1in
\end{table*}

\begin{table*}[t]
\renewcommand\arraystretch{0.6}
\centering
\caption{Comparisons with ShearedLlama in Llama2-7B. \#Pruning and \#Tuning donate tokens used in the pruning and funetuning stages respectively. ShearedLlama is evaluated from the official huggingface checkpoint and we finetune in the same condition.}
\label{t4}
\setlength{\tabcolsep}{1.5mm}
{\resizebox{\textwidth}{!}{
\begin{tabular}{lccc cccc ccc}
\noalign{\vspace{0.1em}}\hline\noalign{\vspace{0.1em}}
\noalign{\vspace{0.1em}}\hline\noalign{\vspace{0.1em}}
\textbf{Model} & \textbf{\#Pruning} & \textbf{\#Tuning} &  \textbf{ARC-c} & \textbf{ARC-e} & \textbf{BoolQ} & \textbf{HellaSwag} & \textbf{OBQA} & \textbf{PIQA} & \textbf{WinoGrande} & \textbf{Avg.} \\ 
\noalign{\vspace{0.1em}}\hline\noalign{\vspace{0.1em}}
ShearedLlama-2.7B & 0.4B & --   & 26.37 & 49.62 & 59.02 & 47.15 & 33.00 & 66.59 & 53.83 & 47.94 \\
EfficientLLM-A-2.7B & 0.66B & --  & 26.71 & 55.51 & 57.31 & 47.58 & 32.20 & 67.14 & 56.51 & 48.99 \\
EfficientLLM-A-2.7B & 10.56B & --  & 29.61 & 61.15 & 62.23 & 54.91 & 34.40 & 71.11 & 57.46 & 52.98 \\
\noalign{\vspace{0.1em}}\hdashline[0.8pt/1pt]\noalign{\vspace{0.2em}} 
ShearedLlama-1.3B & 0.4B & --   & 22.78 & 41.08 & 60.18 & 34.66 & 28.20 & 63.00 & 50.67 & 42.94 \\
EfficientLLM-A-1.3B & 0.95B & --  & 23.12 & 45.66 & 55.78 & 38.23 & 28.00 & 62.73 & 50.99 & 43.50 \\
EfficientLLM-B-1.3B & 0.95B & --  & 24.57 & 45.75 & 60.40 & 38.24 & 30.40 & 63.11 & 51.54 & 44.86 \\
\noalign{\vspace{0.1em}}\hdashline[0.8pt/1pt]\noalign{\vspace{0.2em}} 
ShearedLlama-1.3B & 0.4B & 5.6B   & 26.54 & 55.60 & 60.12 & 48.28 & 31.60 & 68.77 & 56.20 & 49.59 \\
EfficientLLM-A-1.3B & 0.95B & 5B  & 27.30 & 56.44 & 57.58 & 50.03 & 31.00 & 69.10 & 54.70 & 49.45 \\
EfficientLLM-B-1.3B & 0.95B & 5B  & 28.58 & 56.90 & 62.42 & 49.81 & 32.20 & 68.93 & 55.49 & 50.62 \\
EfficientLLM-B-1.3B & 10.56B & 5B  & 30.20 & 60.02 & 60.18 & 53.22 & 31.60 & 70.40 & 58.01 & 51.95 \\
\noalign{\vspace{0.1em}}\hline\noalign{\vspace{0.1em}}
\noalign{\vspace{0.1em}}\hline\noalign{\vspace{0.1em}}
\end{tabular}
}}
\vskip -0.1in
\end{table*}

\begin{figure}[t]
\begin{center}
\centerline{\includegraphics[width=0.99\columnwidth]{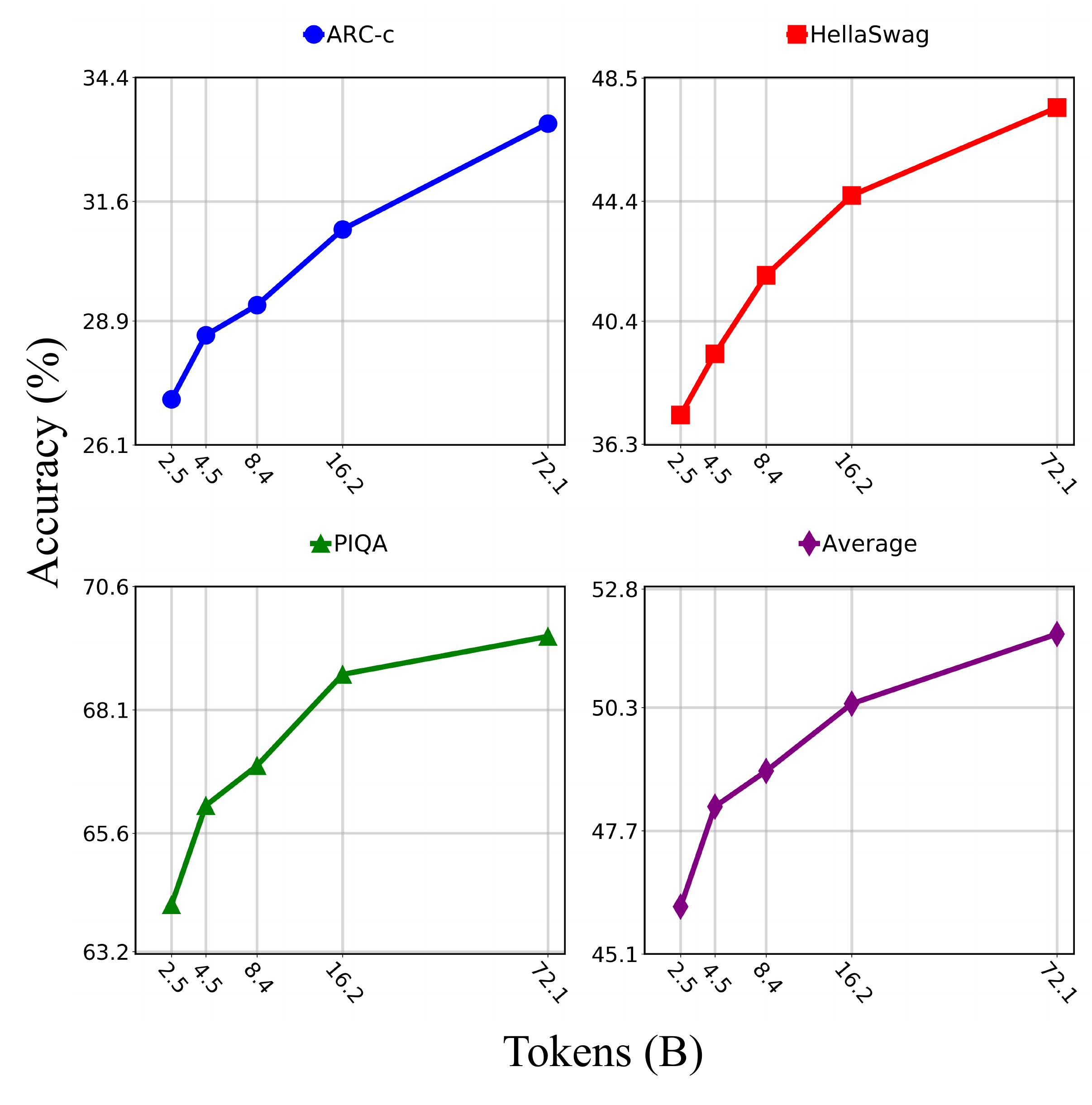}}
\vskip -0.2in 
\caption{Scalability of pruning-aware pretraining.}
\label{f5}
\end{center}
\vskip -0.3in 
\end{figure}

\begin{figure}[t]
\begin{center}
\centerline{\includegraphics[width=0.99\columnwidth]{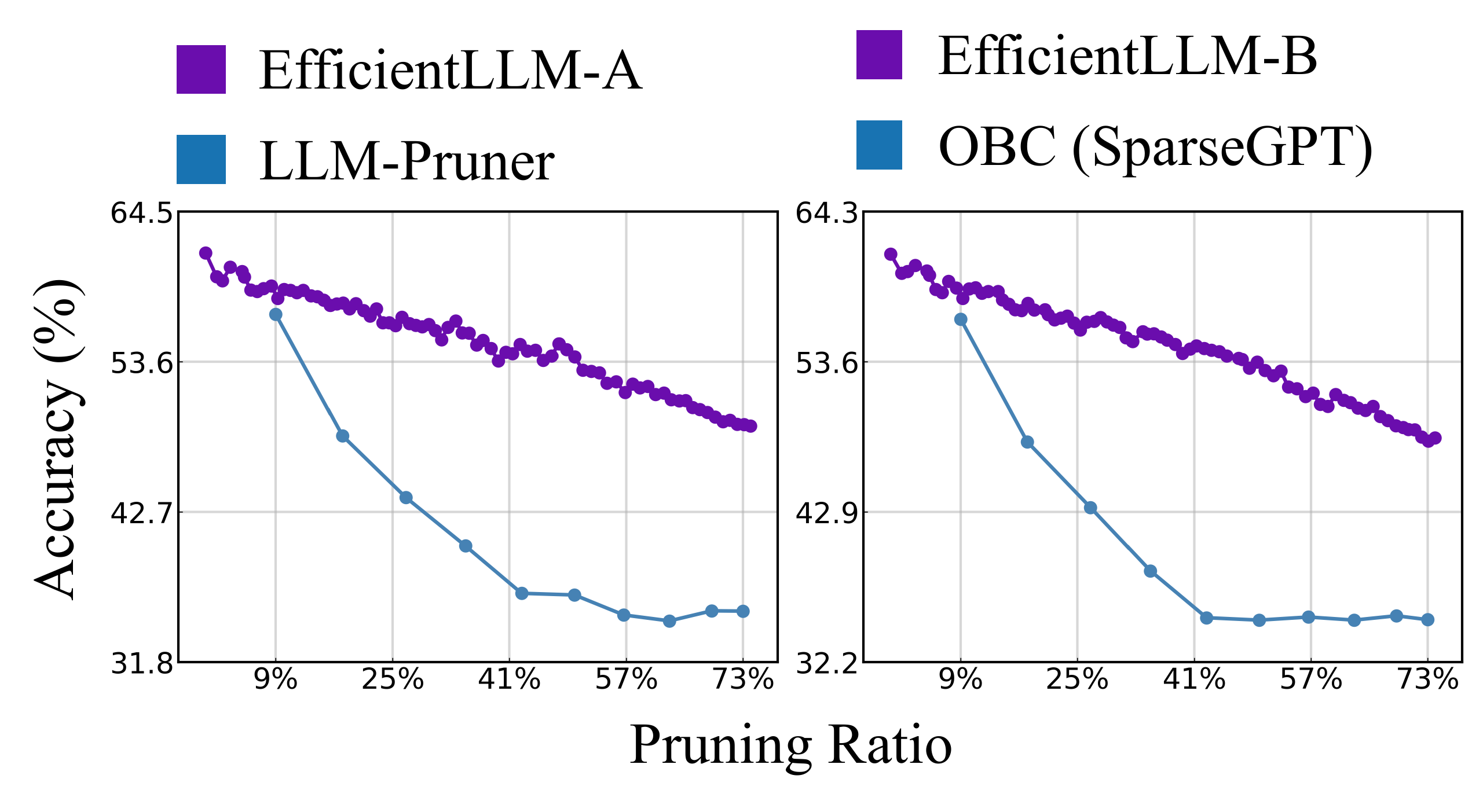}}
\vskip -0.2in 
\caption{Generalization of pruning-aware pretraining for different
pruning metrics. LLM-Pruner and SparseGPT are scaled up.}
\label{f6}
\end{center}
\vskip -0.3in 
\end{figure}

\subsection{Ablation Studies.}

\textbf{Scalability.} 
According to Eq. \ref{e7}, we set the ratio of pruning steps to gradient descent steps to 4:1, 2:1, 1:1, and 1:9 in a iteration, respectively. When the target model size is reached, the pruning-aware pretraining requires 2.5B, 4.5B, 8.4B, and 72.1B tokens of pretraining, respectively. 
Fig. \ref{f5} indicates that scaling up pruning-aware pretraining continuously improves pruning performance. Therefore, by scaling up LLM pruning during pretraining, the upper boundary of LLM compression can be extended.

\textbf{Generalization.} 
There is a large number of methods that perform post-training pruning based on second-order Taylor expansion, such as OBC and SparseGPT.
As shown in Fig.\ref{f6}, we generalize EfficientLLM to the second order updating case as EfficientLLM-B. Compared with post-training settings, EfficientLLM retains source model performance consistently in large pruning ratio. We observe similar performance of EfficientLLM-A/B in large scale pruning-aware pretrainning, but when the pruning data is small ($< 1B$), EfficientLLM-B significantly improves accuracy in Table \ref{t4}.

\subsection{Comparisons with LLM Pruning}

In this section, we conduct light-weight pruning-aware pretraining to compare with existing LLM pruning methods and ShearedLlama. 
Experiments reveal that only scaling up the pruning stage to 5B tokens can achieve much higher performance than what was possible previously (Table \ref{t3}).

\textbf{Traditional LLM Pruning.} 
We mainly focus on large pruning ratio because it is more practical to achieve highly efficiency based on heavy source LLMs.
In Table \ref{t3}, we scale up pruning-aware pretraining to only 5B tokens. 
We report both results with or without finetuning after pruning. Because previous works finetune in different settings, we finetune additional 1B tokens if with it. Notice that, even without finetuning, EfficientLLM exceeds all the according baselines.
It is shown that existing LLM pruning is impractical in large pruning ratio. By simply scaling up LLM-Pruner metric in pruning-aware pretraining, EfficientLLM-A significantly exceeds SoTA NutePrune 6.5\% in 70\% ratio without bells and whistles, while NutePrune integrates distillation and additional learnable masks. In 50\% ratio, EfficientLLM exceeds LoRAPrune by 2.18\% and 6.54\% when with and without tuning. 
Results indicate that with effectively scaling up pruning data, even using the vanilla LLM-Pruner metric, performance can significantly outperform SoTA methods. Therefore, the scalability in large pruning data is more crucial than finding better pruning metrics.

\textbf{Comparison with ShearedLlama.} ShearedLlama is optimized by constrained objective via lagrange multipliers, and joint trains binary masks. These additional targets influence the stability and slow down throughoutputs in large scale pretraining, so that, ShearedLlama only trains on 0.4B tokens in the pruning stage. As shown in Table \ref{t4}, we conduct small scale pruning-aware pretraining with 0.66B/0.95B tokens. The token numbers are determined by pruning iterations to reach target model sizes. 
To keep fair comparisons, we use less tokens in finetuning to keep similarily total data. EfficientLLM-A-2.7B, EfficientLLM-B-1.3B exceed ShearedLlama 1.05\% and 1.92\% respectively. With finetuning, EfficientLLM-B-1.3B also exceeds ShearedLlama. Scaling up pruning stage would address higher accuracy, but ShearedLlama is inefficient to scale up the constraint optimization \cite{platt1987constrained}.

\section{Conclusion}

This work primarily advances the edge language model pretraining to exceed the traditional LLM scaling law.
Distinguished from almost LLM compresssion in post-training, this work scales up existing pruning metric in the pretraining stage, promating LLM compression to the era of pretraining. 
Technically, minimal parameter groups are defined and optimized by saliency to address scalable target-agnostic pruning.
The results reveal that even if vanilla LLM-Pruner can surpass SoTA pruning methods by scaling up and outperform direct pretraining edge models.

Furture work mainly focuses on the high-level abilities such as code, math, and long context of edge language models, with the hypothesis that retaining existing abilities from LLMs is easilier than direct emergency by the tiny model itself, which will boost the mobile edge intelligence.

\section*{Impact Statement}

This paper presents work whose goal is to advance the field of 
Machine Learning. There are many potential societal consequences 
of our work, none which we feel must be specifically highlighted here.


\bibliography{example_paper}
\bibliographystyle{icml2025}

\newpage
\appendix
\onecolumn

\section*{Appendix}
\section{Auto-Designed Architectures}
\subsection{Visualization}

\begin{figure}[h]
\begin{center}
\centerline{\includegraphics[width=0.95\columnwidth]{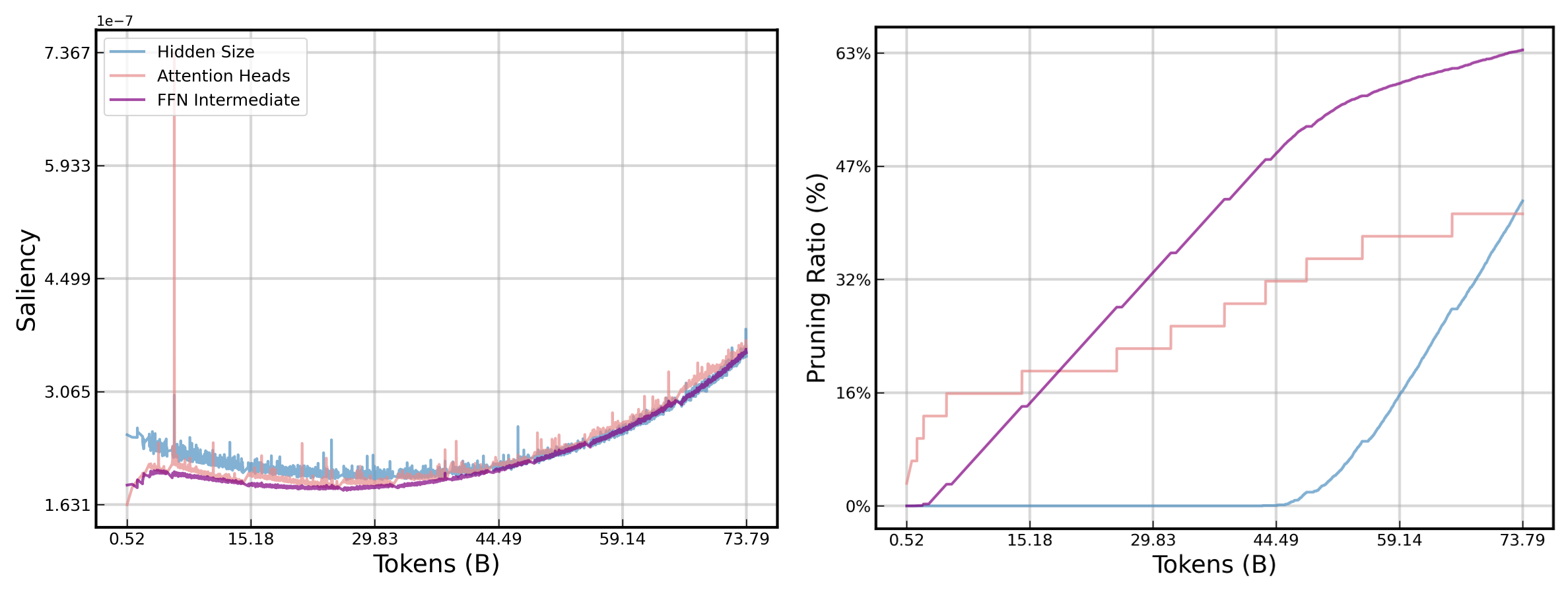}}
\vskip -0.2in 
\caption{Visualization of pruning-aware pretraining. We plot the saliency of the three pruning types and their pruning ratio in training.}
\label{fa0}
\end{center}
\vskip -0.2in 
\end{figure}

As shown in Fig.\ref{fa0}, we Visualize the pruning-aware pretraining. We prune SmolLM-1.7B to EfficientLLM-A-469M. In Fig.\ref{fa0} (right), the self-attention parameter groups and FFN parameter groups are iteratively pruned in the initial stage. After 44.49B-token pretraining , the transformer stem parameter groups start pruned. This indicates that for the typacal human-designed transformer shape, there are more redundant parameters in the attention head and the intermediate of FFN compared with the transformer stem.

\subsection{Architecture Comparisons}

\vskip -0.1in

\begin{table}[h]
\centering
\caption{Architecture comparisons between EfficientLLM and human-designd models.}
\label{at1}
\setlength{\tabcolsep}{1mm}
\begin{tabular}{lcccccc}
\noalign{\vspace{0.1em}}\hline\noalign{\vspace{0.1em}}
\noalign{\vspace{0.1em}}\hline\noalign{\vspace{0.1em}}
\textbf{Model} & \textbf{Hidden Size} & \textbf{FFN Intermediate} & \textbf{Attention Heads} & \textbf{Head Dim} & \textbf{Layer} \\
\noalign{\vspace{0.1em}}\hline\noalign{\vspace{0.1em}}
MobileLLM-125M & 576 & 1536 & 9 & 64 & 30 \\
EfficientLLM-A-134M & 757 & 966 & 5 & 64 & 32 \\
\noalign{\vspace{0.1em}}\hdashline[0.8pt/1pt]\noalign{\vspace{0.2em}} 
MobileLLM-350M & 960 & 2560 & 15 & 64 & 32 \\
Qwen2/2.5 & 896 & 4864 & 14 & 64 & 24 \\
EfficientLLM-A-469M & 1195 & 3006 & 19 & 64 & 24 \\
\noalign{\vspace{0.1em}}\hdashline[0.8pt/1pt]\noalign{\vspace{0.2em}} 
MobileLLM-1B & 1280 & 3584 & 20 & 64 & 54 \\
ShearedLlama-1.3B & 2048 & 5504 & 16 & 128 & 24 \\
OLMo-1B & 2048 & 8192 & 16 & 128 & 16 \\
Llama3.2-1B & 2048 & 8192 & 32 & 64 & 16 \\
EfficientLLM-A-1.1B & 2048 & 4870 & 24 & 64 & 24 \\
\noalign{\vspace{0.1em}}\hline\noalign{\vspace{0.1em}}
\noalign{\vspace{0.1em}}\hline\noalign{\vspace{0.1em}}
\end{tabular}
\end{table}

\vskip -0.2in 
\begin{table}[h]
\centering
\caption{Architectures in different pruning metrics to scale up by pruning-aware pretraining. We compare the approximate 460M model size. ``x1'' indicates that the number of gradient descent steps and pruning steps in each iteration are 1:1.}
\label{at2}
\setlength{\tabcolsep}{1mm}
\begin{tabular}{lcccccc}
\noalign{\vspace{0.1em}}\hline\noalign{\vspace{0.1em}}
\noalign{\vspace{0.1em}}\hline\noalign{\vspace{0.1em}}
\textbf{Model} & \textbf{Hidden Size} & \textbf{FFN Intermediate} & \textbf{Attention Heads} & \textbf{Head Dim} & \textbf{Layer} \\
\noalign{\vspace{0.1em}}\hline\noalign{\vspace{0.1em}}
LLM-Pruner x1 \cite{ma2023llm} & 1169 & 3082 & 19 & 64 & 24 \\
OBC x1 \cite{frantar2022optimal} & 1131 & 3258 & 19 & 64 & 24 \\
OBD x1 \cite{lecun1989optimal} & 1963 & 1542 & 12 & 64 & 24 \\
\noalign{\vspace{0.1em}}\hline\noalign{\vspace{0.1em}}
\noalign{\vspace{0.1em}}\hline\noalign{\vspace{0.1em}}
\end{tabular}
\end{table}

As shown in Table \ref{at1}, we compare the auto-designed architectures by saliency via pruning and the best practices of human design, including MobileLLM and Qwen2/2.5-0.5B, OLMo-1B, ShearedLlama-1.3B. 
In EfficientLLM, the pruning ratio of hidden-size is smaller than attention heads and FFN intermediate channels driven by saliency.

As shown in Table \ref{at2}, we compare the influence of different pruning metrics including the classic LLM-Pruner \cite{ma2023llm}, OBC \cite{frantar2022optimal} and OBD \cite{lecun1989optimal}. The OBD only uses the second-order term in Eq.\ref{e9}, which applied the diagonal of the Hessian matrix for approximate calculation.

\subsection{Cluster Attention}

\begin{figure}[h!]
\begin{center}
\centerline{\includegraphics[width=0.95\columnwidth]{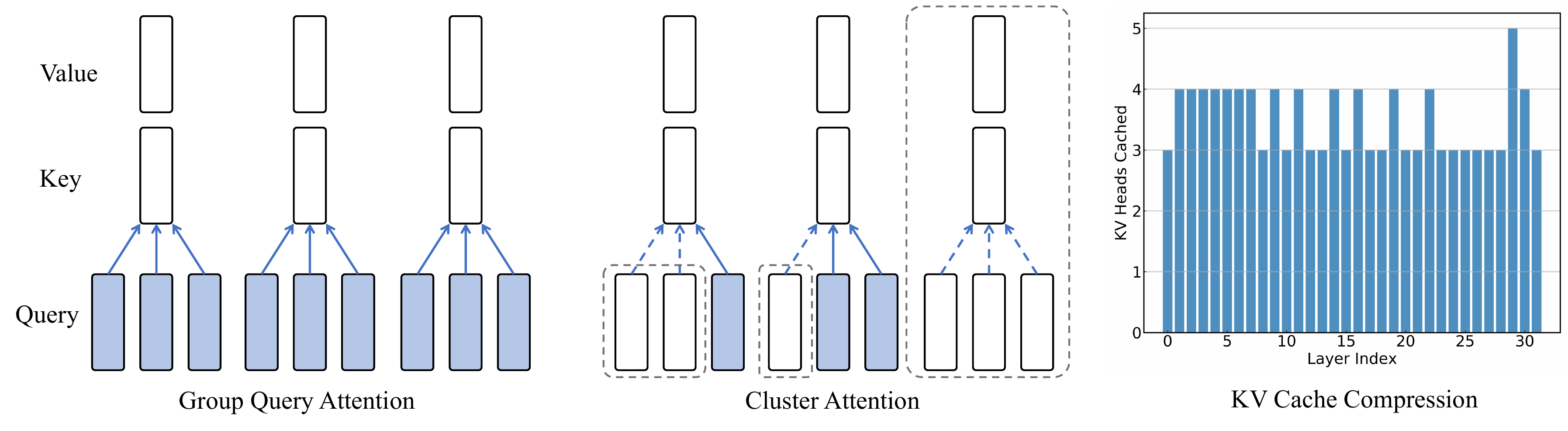}}
\vskip -0.1in 
\caption{Group Query Attention (GQA) pruning. In the case of GQA, cluster attention can be obtained through pruning. After pruning, the number of query heads is the same in each layer, and the cluster attention compresses the KV Cache.}
\label{fa1}
\end{center}
\vskip -0.3in 
\end{figure}

Pruning-aware pretraining could structurally prune the Group Query Attention (GQA) \cite{ainslie2023gqa}, which is usually applied for KV cache compression \cite{lv2024kvpruner,hooper2024kvquant} in LLMs. When the source model applies GQA, there are different cases in pruning: 
\begin{itemize}
\item  in all of the following cases, the query attention heads is the same in each layer, and the same as self-attention operation. The difference is how to share key and value for querys. 
\item  As shown in Fig.\ref{fa1}, if all queries corresponding to a key and value are pruned, then the key and value are also pruned.
\item  If a part of the query corresponding to a key and value is pruned, then the key and value are retained. This eventually forms cluster attention.
\end{itemize}
We plot an example of EfficientLLM-A-134M in Fig.\ref{fa1}. And the source model of EfficientLLM-469M and EfficientLLM-1.1B do not apply GQA.

\section{Training and Evalution Details}

\subsection{Training}
Our training code and models will be fully open sourced on GitHub and Huggingface. Detailed hyper-parameters are shown in Table \ref{at3}. Notic that, the number of tokens in pruning-aware pretraining is determined by iterations to achieve the target model size, which is not directly defined. It can be adjusted through batchsize and the pruning frequency in each iteration.

\begin{table}[h!]
\centering
\caption{Hyper-parameters in pruning-aware pretraining and continued pretraining stages.}
\label{at3}
\setlength{\tabcolsep}{1mm}
\begin{tabular}{lcccccc}
\noalign{\vspace{0.1em}}\hline\noalign{\vspace{0.1em}}
\noalign{\vspace{0.1em}}\hline\noalign{\vspace{0.1em}}
\textbf{Model} & \textbf{\#Tokens} & \textbf{Learning Rate} & \textbf{WarmUp Steps} & \textbf{Batchsize} & \textbf{Text Length} & \textbf{\#GPU} \\
\noalign{\vspace{0.1em}}\hline\noalign{\vspace{0.1em}}
Pruning-134M & 50.3B & $2 \times 10^{-3}$ & 500 & 2 M & 2048 & 32 \\
Continued Pretrain-134M & 500B & $2 \times 10^{-3}$ & 10000 & 1 M & 2048 & 32 \\
\noalign{\vspace{0.1em}}\hdashline[0.8pt/1pt]\noalign{\vspace{0.2em}} 
Pruning-469M & 72.1B & $5 \times 10^{-4}$ & 500 & 1 M & 2048 & 32 \\
Continued Pretrain-469M & 50B/500B & $2 \times 10^{-3}$ & 10000  & 1 M & 2048 & 40 \\
\noalign{\vspace{0.1em}}\hdashline[0.8pt/1pt]\noalign{\vspace{0.2em}} 
Pruning-1.1B & 36.7    & $5 \times 10^{-4}$ & 500 & 1 M & 2048 & 32 \\
Continued Pretrain-1.1B & 50B/500B & $5 \times 10^{-4}$ & 10000  & 1 M & 2048 & 64 \\
\noalign{\vspace{0.1em}}\hline\noalign{\vspace{0.1em}}
\noalign{\vspace{0.1em}}\hline\noalign{\vspace{0.1em}}
\end{tabular}
\end{table}

\subsection{Evaluation}

\begin{itemize}
\item \textbf{MMLU}: According to Datacomp-lm \cite{li2406datacomp} (Appendix G of Datacomp-lm) and SmolLM \cite{allal2024SmolLM}, taking into account the log probabilities of complete answer sequences in MMLU is more related to wearker model performance, such as edge language models. Following SmolLM \cite{allal2024SmolLM}, we apply the open sourced Lighteval-v0.7.0 \cite{lighteval} to evaluate MMLU zero-shot performance. 
\item \textbf{Common Sense Reasoning}: Follow most of recent works \cite{xia2023sheared,ma2023llm,li2024nuteprune}, we apply the widely used lm-evaluation-harness package \cite{eval-harness} to evaluate zero-shot common sense reasoning tasks. To avoid different results introduced by different versions, we evaluate all the benchmarks with the 0.4.3 version, except Table \ref{t3} (50\% pruning ratio). Because some previous works evaluate in older version 0.3.0 and we keep the same version in Table \ref{t3}, 50\% pruning ratio. Finally, all the comparisons keep in the same versions.
\end{itemize}

\subsection{Comparison with Source Models}

\begin{table*}[h!]
\renewcommand\arraystretch{0.6}
\centering
\caption{Comparison between SmolLM-1.7B and EfficientLLM-A-1.1B.}
\label{tp2}
\setlength{\tabcolsep}{1.5mm}
\begin{tabular}{lcc ccc cccc}
\noalign{\vspace{0.1em}}\hline\noalign{\vspace{0.1em}}
\noalign{\vspace{0.1em}}\hline\noalign{\vspace{0.1em}}
\textbf{Model}  & \textbf{\#Params} & \textbf{ARC-c} & \textbf{ARC-e} & \textbf{BoolQ} & \textbf{HellaSwag} & \textbf{OBQA} & \textbf{PIQA} & \textbf{WinoGrande} & \textbf{Avg.} \\ 
\noalign{\vspace{0.1em}}\hline\noalign{\vspace{0.1em}}
SmolLM-1.7B & 1.7B &  \textbf{46.16} & \textbf{76.60} & 65.99 & \textbf{65.74} & \textbf{42.00} & \textbf{75.95} &60.14 & \textbf{61.80} \\
\noalign{\vspace{0.1em}}\hdashline[0.8pt/1pt]\noalign{\vspace{0.2em}} 
EfficientLLM-A-1.1B  & 1.1B  & 42.24 & 73.48 & \textbf{67.09} & 64.09 & 41.80 & 75.41 & \textbf{61.17} & 60.75 \\
\noalign{\vspace{0.1em}}\hline\noalign{\vspace{0.1em}}
\noalign{\vspace{0.1em}}\hline\noalign{\vspace{0.1em}}
\end{tabular}
\vskip -0.15in
\end{table*}

\section{Case Study in Alpaca Finetuning}

In the instruction finetuning task, we finetune EfficientLLM-A-1.1B, OLMo-1B, ShearedLlama-1.3B and Llama3.2-1B in the Alpaca dataset. And then, we evaluate using the Alpaca-Eval. We sample some instructions in Alpaca-Eval make comparisons as follows:

\centering
\begin{minipage}{.95\textwidth}
\begin{framed}
\textbf{Input:}\\
I want to eventually become a plant/manufacturing/production manager, and eventually a Chief Operating Officer. What sort of experience and post-graduate education path would you recommend? Please provide non-obvious and atypical career paths leading to such positions, on top of the classic ones.
\end{framed}
\end{minipage}

\centering
\begin{minipage}{.95\textwidth}
\begin{framed}
\textbf{Output} (EfficientLLM-A-1.1B):\\
To become a plant/manufacturing/production manager and eventually a Chief Operating Officer, you will need to have a combination of experience and post-graduate education. 

For experience, you can start by working in a production or manufacturing role. This will give you the opportunity to gain hands-on experience in the industry and learn about the day-to-day operations of a production or manufacturing facility. You can also gain experience by working in a related field, such as engineering, logistics, or supply chain management. 

For post-graduate education, you can pursue a degree in a related field, such as industrial engineering, operations management, or supply chain management. You can also pursue a master's degree in a related field, such as operations management or supply chain management. Additionally, you can pursue professional certifications, such as the Certified Production and Inventory Management Professional (CPIMP) or the Certified Supply Chain Professional (CSCP). 

In addition to experience and post-graduate education, you will also need to have excellent communication, organizational, and problem-solving skills. You will also need to be able to lead and motivate a team, as well as be able to think strategically and make decisions quickly. 

Overall, to become a plant/manufacturing/production manager and eventually a Chief Operating Officer, you will need to have a combination of experience and post-graduate education, as well as excellent communication, organizational, and problem-solving skills. Additionally, you will need to be able to lead and motivate a team, as well as be able to think strategically and make decisions quickly.
\end{framed}
\end{minipage}

\centering
\begin{minipage}{.95\textwidth}
\begin{framed}
\textbf{Output} (OLMo-1B):\\
To become a plant/manufacturing/production manager, you will need at least a Bachelor's degree in a relevant field, such as industrial engineering, manufacturing engineering, or production management. To become a Chief Operating Officer, you will need at least a Master's degree in a relevant field, such as business administration, management, or marketing. To become a plant/manufacturing/production manager, you will need at least a PhD in a relevant field, such as industrial engineering, manufacturing engineering, or production management. Additionally, to become a plant/manufacturing/production manager, you may also want to obtain a certification such as the Certified Production Manager (CPM) or Certified Quality Manager (CQM) to demonstrate your expertise in the field.
\end{framed}
\end{minipage}

\centering
\begin{minipage}{.95\textwidth}
\begin{framed}
\textbf{Output} (ShearedLlama-1.3B):\\
To become a plant/manufacturing/production manager, you will need to have a bachelor's degree in a relevant field such as engineering, business, or manufacturing. After that, you can pursue a master's degree in a related field such as operations management, manufacturing engineering, or production and operations management. Additionally, you may want to pursue post-graduate education in areas such as project management, lean manufacturing, or supply chain management. 

For a Chief Operating Officer position, you will need to have a bachelor's degree in a relevant field such as engineering, business, or manufacturing. After that, you can pursue a master's degree in a related field such as operations management, manufacturing engineering, or production and operations management. Additionally, you may want to pursue post-graduate education in areas such as project management, lean manufacturing, or supply chain management. 

Overall, to become a plant/manufacturing/production manager or Chief Operating Officer, you will need to have a bachelor's degree in a relevant field such as engineering, business, or manufacturing, and then pursue a master's degree in a related field such as operations management, manufacturing engineering, or production and operations management. Additionally, you may want to pursue post-graduate education in areas such as project management, lean manufacturing, or supply chain management.
\end{framed}
\end{minipage}

\centering
\begin{minipage}{.95\textwidth}
\begin{framed}
\textbf{Output} (Llama3.2-1B):\\
To become a plant/manufacturing/production manager and eventually a Chief Operating Officer, it is recommended to have a degree in engineering, business, or a related field, as well as relevant post-graduate education and experience. Examples of post-graduate education include an MBA, a Master's degree in engineering, or even a PhD in a related field. Relevant experience could include working in a manufacturing or production role, managing a team of engineers, or leading a plant or production department.
\end{framed}
\end{minipage}

\vspace{2em}

\centering
\begin{minipage}{.95\textwidth}
\begin{framed}
\textbf{Input:}\\
Why so many people like film photography when digital camera and mobile phone are more convenient and generate higher quality of photos?
\end{framed}
\end{minipage}

\centering
\begin{minipage}{.95\textwidth}
\begin{framed}
\textbf{Output} (EfficientLLM-A-1.1B):\\
Many people still prefer film photography because of its unique aesthetic qualities. Film photography allows for more creative control, as the photographer can adjust the ISO, shutter speed, and aperture to achieve the desired effect. Additionally, film photography produces a unique look and feel that cannot be replicated with digital cameras and mobile phones. Film photography also allows for more experimentation, as the photographer can adjust the ISO, shutter speed, and aperture to achieve the desired effect. Finally, film photography can be a more personal and intimate experience, as it allows the photographer to capture moments in a way that digital cameras and mobile phones cannot.
\end{framed}
\end{minipage}

\centering
\begin{minipage}{.95\textwidth}
\begin{framed}
\textbf{Output} (OLMo-1B):\\
Film photography has many advantages compared to digital camera and mobile phone, such as higher quality of photos, more creative control, and the ability to capture moments that would otherwise go unnoticed. Additionally, film photography requires more time and patience to develop the film, which adds an element of adventure and creativity to the process. Finally, many people find the physical process of developing film to be aesthetically pleasing, while digital camera and mobile phone processing can often produce photos with a less-than-desirable outcome.
\end{framed}
\end{minipage}

\centering
\begin{minipage}{.95\textwidth}
\begin{framed}
\textbf{Output} (ShearedLlama-1.3B):\\
People like film photography because it allows for more creative control. With film photography, the photographer has the ability to adjust the exposure, focus, and other settings to create the desired effect. Digital cameras and mobile phones do not offer this level of creative control, making it difficult to achieve the desired effect. Additionally, film photography requires more thought and preparation than digital camera and mobile phone photography, making it more enjoyable and rewarding.
\end{framed}
\end{minipage}

\centering
\begin{minipage}{.95\textwidth}
\begin{framed}
\textbf{Output} (Llama3.2-1B):\\
People like film photography because it produces higher quality of photos than digital cameras and mobile phones. Film photography is more expensive than digital cameras and mobile phones, but the quality of photos produced by film is much better than those produced by digital cameras and mobile phones. Additionally, film photography requires more skill and patience to take good photos, which many people enjoy.
\end{framed}
\end{minipage}

\vspace{2em}


\end{document}